\documentclass{article}

\usepackage{microtype}
\usepackage{graphicx}
\usepackage{subfigure}
\usepackage{booktabs} % for professional tables

\usepackage{hyperref}

\usepackage[accepted]{icml2025}

\usepackage{amsmath}
\usepackage{amssymb}
\usepackage{mathtools}
\usepackage{amsthm}
\usepackage{soul}
\usepackage{amsmath}
\usepackage{algorithm}

\usepackage{algpseudocode}

\usepackage{xcolor}
\usepackage{framed}

\usepackage[capitalize,noabbrev]{cleveref}

\theoremstyle{plain}
\newtheorem{theorem}{Theorem}[section]

\newtheorem{lemma}[theorem]{Lemma}

\theoremstyle{definition}

\newtheorem{assumption}[theorem]{Assumption}
\theoremstyle{remark}

\usepackage[textsize=tiny]{todonotes}

\begin{document}

\twocolumn[
\icmltitle{Fluid Democracy in Federated Data Aggregation}



\icmlsetsymbol{equal}{*}

\begin{icmlauthorlist}
\icmlauthor{Aditya Vema Reddy Kesari}{yyy}
\icmlauthor{Krishna Reddy Kesari}{comp}
\end{icmlauthorlist}

\icmlaffiliation{yyy}{IIT Bombay, India}
\icmlaffiliation{comp}{Amazon, US [This work does not relate to my position at Amazon]}

\icmlcorrespondingauthor{Aditya}{22b3985@iitb.ac.in}

\icmlkeywords{Machine Learning, ICML}

\vskip 0.3in
]



\printAffiliationsAndNotice{}  % leave blank if no need to mention equal contribution

\begin{abstract}
Federated learning (FL) mechanisms typically require each client to transfer their weights to a central server, irrespective of how useful they are. In order to avoid wasteful data transfer costs from clients to the central server, we propose the use of consensus based protocols to identify a subset of clients with most useful model weights at each data transfer step. First, we explore the application of existing fluid democracy protocols to FL from a performance standpoint, comparing them with traditional one-person-one-vote (also known as \emph{1p1v} or FedAvg). We propose a new fluid democracy protocol named \emph{viscous-retained democracy} that always does better than \emph{1p1v} under the same assumptions as existing fluid democracy protocols while also not allowing for influence accumulation. Secondly, we identify weaknesses of fluid democracy protocols from an adversarial lens in terms of their dependence on topology and/ or number of adversaries required to negatively impact the global model weights. To this effect, we propose an algorithm (FedVRD) that dynamically limits the effect of adversaries while minimizing cost by leveraging the delegation topology.
\end{abstract}

\section{Introduction}
\label{Introduction}

Federated Learning (FL) offers a promising paradigm for training machine learning models by keeping sensitive data localized on client devices and only sharing model updates with a central server for aggregation. This approach prioritizes data privacy, as it allows for the collaborative improvement of a global model without direct access to raw client data. Typically, the central server averages all client weights to arrive at the global weights \cite{fedavg}, which is equivalent to one-person-one-vote (\emph{1p1v}). This requires model weight transfer from all clients to central server at every training step resulting in a high cost overhead. In this paper, our first contribution is the application of fluid democracies to FL so as to leverage their inherent property of reducing number of clients that transfer data to the central server, thereby reducing data transfer cost. 

Fluid democracies \cite{boldi, liqdem1, viscous2} enable a voter to delegate their vote to a more competent neighbor, who then casts a vote on both their behalf. Liquid democracy reduces the number of voters and achieves improved performance over \emph{1p1v} voting under mild conditions. However, it is plagued by the disadvantage of few voters accumulating a high number of votes, which can affect performance if they are misaligned. This makes liquid democracy vulnerable to malicious agents. Viscous democracy was proposed to reduce influence accumulation by decaying the value of a vote with every delegation hop. Our second contribution is to show that viscous democracy achieves sub-\emph{1p1v} performance more often than liquid democracy. Subsequently, we identify that this sub-\emph{1p1v} performance arises due to a topological dependence. Thus, although viscous democracy addresses the issue of misaligned voters to some extent, the topological dependence introduced can be exploited by adversaries.

As our third contribution, to address these limitations, we propose \emph{viscous-retained democracy} that a) achieves higher performance than \emph{1p1v} under same conditions as liquid democracy, b) addresses the above weakness of viscous democracy by reducing adversaries' ability to exploit topological dependencies and c) maintains the beneficial characteristic of viscous democracy by limiting influence aggregation. However, it incurs a higher cost than both liquid and viscous democracies, that stems from a larger number of voters casting their vote. To this effect, we propose an algorithm that leverages viscous-retained democracy coupled with a threshold to solve the dual-optimization problem of cost and robustness by effectively leveraging the delegation topology.

Section \ref{related work} provides an overview of related work. Section \ref{model} introduces the formal theoretical setup, outlining our social choice framework, delegation function and delegation graph. Section \ref{propagation} presents our theoretical results on voting processes. Section \ref{experimental} details the experimental analysis conducted to validate the above theoretical results following which we propose an algorithm to effectively apply our theoretical and experimental findings in FL. Finally, Section \ref{future work} discusses our next steps and potential extensions.

\section{Related Work}
\label{related work}
As this work lies at the intersection of adversarial defense and election mechanisms, we detail related work on both aspects in the context of FL.

\subsection{Adversarial Defense in FL}
Several existing solutions protect against vulnerability of FL to malicious client updates at the central server. \cite{blanchard} introduced robust aggregation strategies like Krum and Trimmed Mean, which selectively filter or disregard outlier updates. Building on this, \cite{yin} further explored Byzantine-tolerant gradient descent, providing theoretical guarantees for robustness. Recognizing the inherent resilience of median-based approaches, subsequent research gravitated towards geometric median variants. \cite{Pillutla} proposed Robust Federated Aggregation (RFA), leveraging the geometric median of local model updates to effectively counter poisoning attacks. Beyond statistical robustness, \cite{Cao} introduced FLTrust, a mechanism that evaluates the trustworthiness of client updates using root of trust to filter suspicious contributions. Zeno \cite{xie2020zeno} prioritized updates that demonstrably improve the global model's performance on a validation set. \cite{li} provides empirical comparisons of various schemes under diverse attack scenarios and data heterogeneity.

Contrary to the above methods, the central server in our proposed framework a) dynamically adapts to provide robustness leveraging information of the delegation topology adopted by the clients and b) accepts data transfer from a subset of clients in the delegation graph elected by the clients themselves, both occurring at every data transfer step. 

\subsection{Client Election Mechanisms in FL}
 FedVote \cite{yue2022fedvote} and FedVoting \cite{fedvoting} utilize plurality voting, treating validation results as votes to determine the optimal global model. Similarly, \cite{sohn2020election} introduce election coding to identify majority opinions during aggregation. DETOX \cite{rajput2019detox} employs a hierarchical aggregation strategy based on majority votes within groups of updates.  Furthermore, election mechanisms are crucial for robustness against attacks during aggregation, as demonstrated by DRACO \cite{chen2018draco} and ByzShield \cite{konstantinidis2021byzshield}, both of which use majority voting to defend against malicious contributions. 
 
 More recent fluid democracy voting techniques such as liquid democracy \cite{liqdem1, liqdem2, liqdem3} offer avenues for voters to delegate their vote to a more informed neighbor so as to increase overall performance. Viscous democracy \cite{boldi, viscous2} builds on it to reduce the extent of influence aggregation in addition to better overall performance in certain cases. In this paper, we present the applicability of fluid democracy protocols as client-side aggregation methods in FL to enable efficient weight transfer from clients to the central server at each training step, leading to cost benefits. In addition, we propose viscous-retained democracy and explore adversarial robustness of viscous and viscous-retained democracy in FL. 
\section{Model}
\label{model}

In this section, we formally define our social choice framework, delegation process, delegation graph and specific graph topologies. \cite{liqdem4, viscous2}. 

\subsection{Social Choice Framework for Federated Learning}

We consider a fundamental epistemic social choice framework involving $n$ voters, denoted as $V = \{v_1, \ldots, v_n\}$, and two possible alternatives, $A = \{a^+, a^-\}$. We posit that $a^+$ represents the objectively correct outcome, which all voters collectively endeavor to select. Each voter $v_i \in V$ possesses a competency level $q_i \in Q$ within the interval $[0, 1]$, signifying the probability that $v_i$ will vote ``correctly'' (i.e., cast a vote for $a^+$). 

Mapping the above terminology to the FL setting, voters refer to clients, $q_i$ is relevance of client $v_i$'s data to the global task at hand, $a^+$ is the set of universal data points across clients relevant to the global task while $a^-$ is the set of universal data points across clients detrimental to the global task. Voting correctly refers to $v_i$ providing model weight updates from data which is a subset of $a^+$.

\subsection{Delegation Process}

The voters are interconnected through an underlying graph $G = (V, E)$, where $E$ is a set of undirected edges illustrating the relationships between voters through which delegations can occur.

Each voter $v_i$ can undertake the following action:
they can delegate $\alpha$ of their vote to a voter in their set of neighbors, $N_G(i) = \{j \in V | (i, j) \in E\}$ and they keep $\beta(1 - \alpha)$ of their vote them supporting $a^+$ with probability $q_i$. The setting of liquid democracy is described by $\alpha=1$ and viscous democracy by $\beta=0$ with $\alpha \in [0, 1]$. In our proposed viscous-retained democracy, we have $\beta=1$ and $\alpha \in [0, 1]$. If a voter chooses to vote directly without delegation, we categorize them as a \textit{guru}.

The delegation process is formalized by a function $d : V \to V$, where $d(v_i) = v_j$ signifies a delegation from voter $v_i$ to voter $v_j$. This delegation can travel transitively across multiple hops until it reaches a guru. The notation $d^*(v_i)$ represents the repeated application of $d(v_i)$ until a guru is reached, effectively identifying $v_i$'s guru. The set of all gurus is denoted as $G(V)$. In our paper, we employ the Upward Delegation process wherein each delegating voter only delegates to a neighbor more competent than themselves, hence no cycles are formed. We chose this delegation process because it is widely used in relevant literature \cite{liqdem4, viscous2} and intuitively aligns with the idea that a voter would like to delegate to a more competent neighbor.

\subsection{Delegation Graph}
The application of a delegation function creates a directed subgraph of $G$, which we refer to as the delegation graph, $D = (V, \{(i, j) \in E \mid d(i) = j\})$. This graph selectively includes directed versions of edges from $E$ only when a delegation relationship exists between the connected nodes.

\subsection{Accuracy}

To ascertain the winner between the two alternatives, we employ weighted plurality voting, a mechanism consistent with May's theorem \cite{May}. The process involves each guru, leveraging their inherent competence, casting their entire accumulated weight in favor of a single alternative. The outcome of the election is decided by the alternative that garners the highest cumulative weight. The likelihood of $a^+$ being chosen as the winning alternative is referred to as accuracy. This accuracy is directly influenced by the established delegations and the individual voter competencies while also exhibiting a dependence on the parameter $\alpha$.

\subsection{Star and Chain Graphs}
We define star and chain graphs by parameters $(s, n_s, c, n_c)$ which establish a delegation graph comprising $s$ \textit{star} components with $n_s$ nodes and and $c$ \textit{chain} components with $n_c$ nodes. Each star component features one guru at its center and $n_s - 1$ delegators, while each chain component contains $n_c - 1$ delegators, with a guru positioned at one end. 
\vspace{-0.3cm}

\section{Vote Propagation}
\label{propagation}

In this section, we analyze liquid and viscous democracy with respect to the \emph{do no harm} property and propose viscous-retained democracy to address the shortfall. The three key theoretical results are
\begin{itemize}
    \item Liquid democracy satisfies \emph{do no harm} property under mild assumptions.
    \item There exist cases where viscous democracy does not satisfy \emph{do no harm} property when liquid democracy does
    \item Viscous-retained democracy satisfies \emph{do no harm} property  under the same assumptions as liquid democracy.
\end{itemize}

\subsection{Do No Harm Property}

We state the \emph{do no harm} (DNH) property \cite{liqdem1} as the following :

For a mechanism \textit{M} and a graph $G$, we define the gain over direct democracy $D$ as 
\textit{}\[\textit{}
\text{gain}(M, G) = P_M(G) - P_D(G).
\]

The objective of the \emph{do no harm} property to never significantly underperform direct voting formally defined as - 
\begin{itemize}
\item A mechanism $M$ satisfies the \emph{do no harm} property if for all $\varepsilon > 0$, there exists $n_1 \in \mathbb{N}$ such that for all social networks $G_n$ on $n \ge n_1$ vertices, $\text{gain}(M, G_n) \ge -\varepsilon$.
\end{itemize}

The DNH property stipulates that any potential loss of delegative voting relative to direct voting must asymptotically approach zero as the graph size tends to infinity. This implies that while small instances might show delegative voting performing worse than direct voting, this disadvantage should vanish in large-scale settings. 

We state that DNH is of utmost importance to us as the gold standard for FL aggregation of weights is FedAvg which is equivalent to direct democracy. Hence, always performing better than direct democracy would imply performing better than FedAvg in the FL setting.

We begin by stating the following lemma \cite{liqdem4} 

\begin{lemma} If $M$ is a mechanism, there is an $\beta \in (0,1)$ and $C: \mathbb{N} \to \mathbb{N}$ with $C(n) \in o(n)$ such that
\begin{align}
    \max\text{-weight}(G_n) \le C(n)  \label{eq:1} \\
    \sum_{i=1}^n \text{weight}_i(G_n) \cdot p_i - \sum_{i=1}^n p_i \ge 2\beta n \label{eq:2}
\end{align}
then $M$ satisfies do no harm property.
\label{lemma1}
\end{lemma}

\vspace{0.2cm}

\begin{assumption} 
\label{assumption}
In all the proofs in this section, we assume that the number of edges $|E|$ in the delegation graph D having n vertices V satisfies $|E| \geq C(n)$ for some $C(n) \in \mathcal{O}(n)$. We also assume that no two voters have equal competencies.

\end{assumption}

\subsection{Liquid Democracy}

In liquid democracy using the upward delegation process, a voter may either delegate their entire vote to a more competent neighbor or vote independently if no such neighbor exists \cite{liqdem1}. The voting weight of a guru is determined by the total number of direct and indirect delegations they receive.

\begin{lemma}
Liquid democracy satisfies \textit{do no harm} property in upward delegation processes, provided that no guru has $\mathcal{O}(n)$ delegators (direct or indirect), and Assumption \ref{assumption} holds.
\end{lemma}

\begin{proof}

From the first condition, Equation \ref{eq:1} holds directly. 

To establish Equation~(2), since delegation happens only to more competent neighbors, the minimum improvement in competence per delegation is $c_{\min} > 0$. By Assumption~4.2, no two voters have equal competence values. Additionally, since the number of delegations is at least $\mathcal{C}(n)$, the cumulative effect of competence differences across all delegations is some constant times n, satisfying Equation \ref{eq:2}.

\end{proof}

\subsection{Viscous Democracy}

 In viscous democracy \cite{boldi}, each hop a delegation traverses by reduces its weight by a constant dampening factor, $\alpha \in [0, 1]$, referred to as the \textit{viscosity}. Consequently, the weight of each voter $v_i$ is defined as:
\[
w_i =
\begin{cases}
    0 & \text{if } v_i \notin G(V) \\
    \sum_{p \in \text{Path}(-,i)} \alpha^{|p|} & \text{otherwise}
\end{cases}
\]
where $\text{Path}(-,i)$ denotes all delegation paths that terminate at $v_i$ and \textit{G(V)} is the set of all gurus. Key implications are:
\begin{itemize}
    \item When $\alpha = 1$, viscous democracy simplifies to the standard liquid democracy model, where no weight is lost during delegation.
    \item For smaller values of $\alpha$, guru weights are directly influenced by the structure of the delegation graph. Specifically, the weight contributions from voters located further away from their guru diminish more significantly compared to those closer to their guru.
    
\end{itemize}

Although viscous democracy has shown advantages over liquid democracy in some upward delegation settings \cite{viscous2}, it falls short of meeting a key criterion for a well-functioning voting system: the \emph{do no harm} property, even under the assumptions used for liquid democracy. We present the following theorem to show the same. 

\vspace{0.5cm}

\setcounter{theorem}{0} % Reset theorem counter
\renewcommand{\thetheorem}{1} % Force theorem number to be 1

\begin{theorem}
Under the upward delegation process, viscous democracy does not satisfy do no harm property under identical assumptions used for liquid democracy.
\end{theorem}

\renewcommand{\thetheorem}{\arabic{theorem}} % Restore default numbering

\begin{proof}

Consider \(n\) voters are divided into two equally sized families of disjoint subgraphs. The first \(n/2\) voters form chains of length~10, each voter in these chains has competence in the interval \((b,c)\); the guru in each chain accrues at most
\[
\sum_{k=0}^9 \alpha^k \;\le\;\frac{1}{1-\alpha}
\]
votes. The remaining \(n/2\) voters form stars of size~10, each leaf and each central guru having competence \(a\) and all other nodes having competence in the range (q,a), so that the total weight of each guru is \(1 + 9\alpha\). Before delegation, the mean competence of the group is at least \(\frac{q + b}{2}\). After viscous delegation, the weighted mean competence is at most
\[
\frac{\bigl(\tfrac{1}{1-\alpha}\bigr)\,c \;+\;(1 + 9\alpha)\,a}
     {\tfrac{1}{1-\alpha} \;+\;(1 + 9\alpha)}.
\]
Concretely, with $\alpha=0.5$, $q=0.39$, $a=0.4$, $b=0.65$, and $c=0.7$, each chain root accumulates at most $1/(1-\alpha)=2$ votes while each star root accumulates $1+9\alpha=5.5$ votes.  Under these weights, the mean competence after delegation falls from above $0.5$ (in direct voting) to at most $0.48$ (in viscous democracy).  By the Condorcet Jury Theorem, as n increases, direct voting thus converges to the correct majority decision with probability tending to 1, whereas viscous delegation converges to the wrong decision with probability tending to 1.  Hence, viscous democracy can strictly worsen collective accuracy and fails the 'do no harm' requirement. This completes the proof.

\end{proof}

We observe that the primary reason such cases arise is the dynamic nature of the total number of votes. As a result, there are scenarios where a larger fraction of high-competence votes is lost during delegation compared to lower-competence votes, leading to a decrease in the overall mean competence. Voters with relatively high competence located at the end of chain-like topologies tend to lose a large fraction of their voting weight, whereas voters with low competence in star-like topologies retain a significant portion of their voting weight. Consequently, each delegation does not necessarily result in a strict increase in competence, as there different amount of loss of votes. Motivated by this observation, we propose a new delegation mechanism: \emph{viscous-retained democracy}.

\subsection{Viscous-retained democracy}
In viscous-retained democracy, similar to viscous democracy above, each hop a delegation traverses reduces its weight by a constant dampening factor, $\alpha \in [0, 1]$. In addition, in viscous-retained democracy, the delegator retains $1-\alpha$ fraction of their vote post delegation. Consequently, the weight of each voter $v_i$ is defined as:
\[
w_i =
\begin{cases}
     \sum_{p \in \text{Path}(-,i)} (1-\alpha)*\alpha^{|p|}& \text{if } v_i \notin G(V) \\
    \sum_{p \in \text{Path}(-,i)} \alpha^{|p|} & \text{otherwise}
\end{cases}
\]
where $\text{Path}(-,i)$ denotes all delegation paths that terminate at $v_i$ and \textit{G(V)} is the set of all gurus. Key implications are:
\begin{itemize}
    \item When $\alpha = 1$, viscous-retained  democracy also simplifies to the standard liquid democracy model, where no weight is lost during delegation.
    \item The number of votes remain constant irrespective of the value of $\alpha$. The weight of gurus remain same as in the case of viscous democracy, thus retaining its ability to reduce accumulation of votes in the hands of misaligned voters.
    
\end{itemize}

We demonstrate that, in contrast to viscous democracy, viscous-retained democracy preserves the \emph{do no harm} property requiring identical conditions as liquid democracy.

\vspace{0.5cm}
\begin{theorem}
Viscous-retained democracy satisfies the \textit{Do No Harm} property in upward delegation processes, provided that no guru has $\mathcal{O}(n)$ delegators (direct or indirect), and Assumption \ref{assumption} holds.
\end{theorem}

\begin{proof} From the first condition, Equation \ref{eq:1} holds directly. 

To establish Equation \ref{eq:2}, since delegation happens only to more competent neighbors, the minimum competence difference between any delegator and their chosen guru is $c_{\min} > 0$. Let $m$ be the length of the longest path in the delegation graph $D$. Since no guru has $\mathcal{O}(n)$ delegators, $m$ is a constant. Thus, $\alpha^m > 0$. Hence, the minimum competence increase per delegation is $c_{\min} \cdot \alpha^m$. By Assumption \ref{assumption}, no two voters have equal competence values. Additionally, since the number of delegations is at least $\mathcal{C}(n)$, the cumulative effect of competence differences across all delegations is some constant times n, satisfying Equation \ref{eq:2}.
\end{proof}

We motivate the use of viscous-retained democracy by highlighting its improved robustness against malicious agents compared to standard viscous democracy. In  viscous-retained democracy, the most influential voters retain the same voting power as viscous democracy, while the overall vote mass in the system is increased. This enhancement not only increases robustness but also ensures that \emph{do no harm} property is satisfied whenever it holds for liquid democracy.

\section{Experimental Analysis}
\label{experimental}

In networks composed entirely of star and chain topologies, viscous democracy can be viewed as a special case of viscous-retained democracy, specifically the case in which only voters with weight greater than 1 are permitted to vote. In FL scenarios, where communication costs are significant, minimizing the number of active voters is essential. However, reducing the number of voters increases the system's vulnerability to adversarial attacks. To address this trade-off, we propose a hybrid model that blends viscous and viscous-retained democracy. This hybrid model introduces a tunable threshold parameter $\tau \in (0,1)$: lowering $\tau$ increases the number of voters and thereby improves robustness, while higher values reduce communication overhead by limiting participation.

We consider a network $n$ with 40 nodes such that $n_c$ is 10 and $c$ is 4, wherein the guru is positioned at one end of each and $\alpha$ = 0.5. In order to ascertain the clients that will send their model weights to the central server, we pick a threshold $\tau$. The clients who have vote share higher than $\tau$ after all clients vote end up transmitting their model weights to the central server. 

Each transfer of model weights to the central server incurs a cost $c$. The total data transfer cost $c_{total}$ is the product of cost $c$ and the number of clients that are elected to send model weights to the central server. 

We now consider an adversary with a fixed budget $c_{\text{adv}}$. The number of adversarial clients that can actively transfer data, denoted $n_{\text{adv}}$, is determined by the integer division of $c_{\text{adv}}$ by the cost per client $c$. The adversary strategically arranges its $n_{\text{advtot}}$ clients in a topology that maximizes their influence. We note that while the adversary deploys a total of $n_{\text{advtot}}$ agents, only the $n_{\text{adv}}$ clients are capable of casting votes i.e., they are the only ones to accumulate votes above the specified threshold.

% \begin{algorithm}[h!]
% \caption{\textsc{FedVD}}
% \label{alg:FedVD}
% \begin{algorithmic}[1]    % <-- (1) enable line numbers

%   \Statex \textbf{Input:} $w^0 \leftarrow$ random initialization
%   \Statex \textbf{Server:}
%   \For{Iteration $t \leftarrow 1$ to $T$}
%     \State Broadcast $w^{t-1}$ to all clients
%     \State Receive delegations, compute votes $v_i^t$ and optimal $\tau$
%     \State Receive updates from clients with $v_i^t>\tau$
%     \State $w^t_n \leftarrow \dfrac{\sum_i v_i^t\,w^t_{i,n}}{\sum_i v_i^t}$
%   \EndFor

%   \Statex \textbf{Client:}
%       % <-- (2) reset numbering here
%   \For{Client $i$}
%     \State Receive $w^{t-1}$ and do local training  
%            $w^t_i \gets w^{t-1}-r_{t-1}\nabla\ell_i(w^{t-1};\Dcal_i)$
%     \State Compute similarity  
%            $s^t_i \gets \frac{\langle w^t_i,w^{t-1}\rangle}{\|w^t_i\|\;\|w^{t-1}\|}$
%     \State Delegate/vote and send decision to server
%     \If{server requests weights}
%       \State send $w^t_i$
%     \EndIf
%   \EndFor

% \end{algorithmic}
% \end{algorithm}

\begin{algorithm}[h!]
\caption{\textsc{FedVRD}}
\label{alg:FedVD}
\begin{algorithmic}[1] % [1] tells it to number lines
\State \textbf{Input:} $w^0 \leftarrow$ random initialization
\State \textbf{Server:}
\For{Iteration $t \leftarrow 1$ to $T$}
    \State Broadcast $w^{t-1}$ to all clients
    \State Receive delegations, compute votes $v_i^t$ and optimal $\tau$
    \State Receive updates from clients with $v_i^t>\tau$
    \State $w^t_n \leftarrow \dfrac{\sum_i v_i^t\,w^t_{i,n}}{\sum_i v_i^t}$
\EndFor

\State \textbf{Client:}
\For{Client $i$}
    \State Receive $w^{t-1}$ and do local training $w^t_i \gets w^{t-1}-r_{t-1}\nabla\ell_i(w^{t-1};\mathcal{D}_i)$
    \State Compute similarity $s^t_i \gets \frac{\langle w^t_i,w^{t-1}\rangle}{\|w^t_i\|\;\|w^{t-1}\|}$
    \State Delegate/vote and send decision to server
    \If{server requests weights}
      \State send $w^t_i$
    \EndIf
\EndFor

\end{algorithmic}
\end{algorithm}

% \begin{algorithm}[h!]
% \caption{\textsc{FedVRD}}
% \label{alg:FedVD}
% \begin{algorithmic}[1] % No [1] needed here because \State is redefined to handle numbering
% \Statex \textbf{Input:} $w^0 \leftarrow$ random initialization
% \Statex \textbf{Server:}
% \For{Iteration $t \leftarrow 1$ to $T$}
%     \State Broadcast $w^{t-1}$ to all clients
%     \State Receive delegations, compute votes $v_i^t$ and optimal $\tau$
%     \State Receive updates from clients with $v_i^t>\tau$
%     \State $w^t_n \leftarrow \dfrac{\sum_i v_i^t\,w^t_{i,n}}{\sum_i v_i^t}$
% \EndFor

% % To ensure continuous numbering, treat "Client:" as a numbered State
% \State \textbf{Client:}
% % Alternatively, if you want "Client:" unnumbered and not indented, just use:
% % \textbf{Client:} % No \State, so no number. But indentation will be lost.

% \For{Client $i$}
%     \State Receive $w^{t-1}$ and do local training $w^t_i \gets w^{t-1}-r_{t-1}\nabla\ell_i(w^{t-1};\mathcal{D}_i)$
%     \State Compute similarity $s^t_i \gets \frac{\langle w^t_i,w^{t-1}\rangle}{\|w^t_i\|\;\|w^{t-1}\|}$
%     \State Delegate/vote and send decision to server
%     \If{server requests weights}
%       \State send $w^t_i$
%     \EndIf
% \EndFor

% \end{algorithmic}
% \end{algorithm}

We present our adversarial robustness analysis based on the optimal network topology an adversary might construct to attack the voting network of size $n$. For example, in the case of $\tau = 1$, which corresponds to standard viscous democracy, wherein only four agents cast votes, each accumulating slightly less than two votes. Therefore, the adversary would need to secure at least eight votes to gain a majority. 

\begin{figure}[h!] % The figure environment is for floats (images, tables)
    \centering % Centers the image horizontally
    \includegraphics[width=\columnwidth]{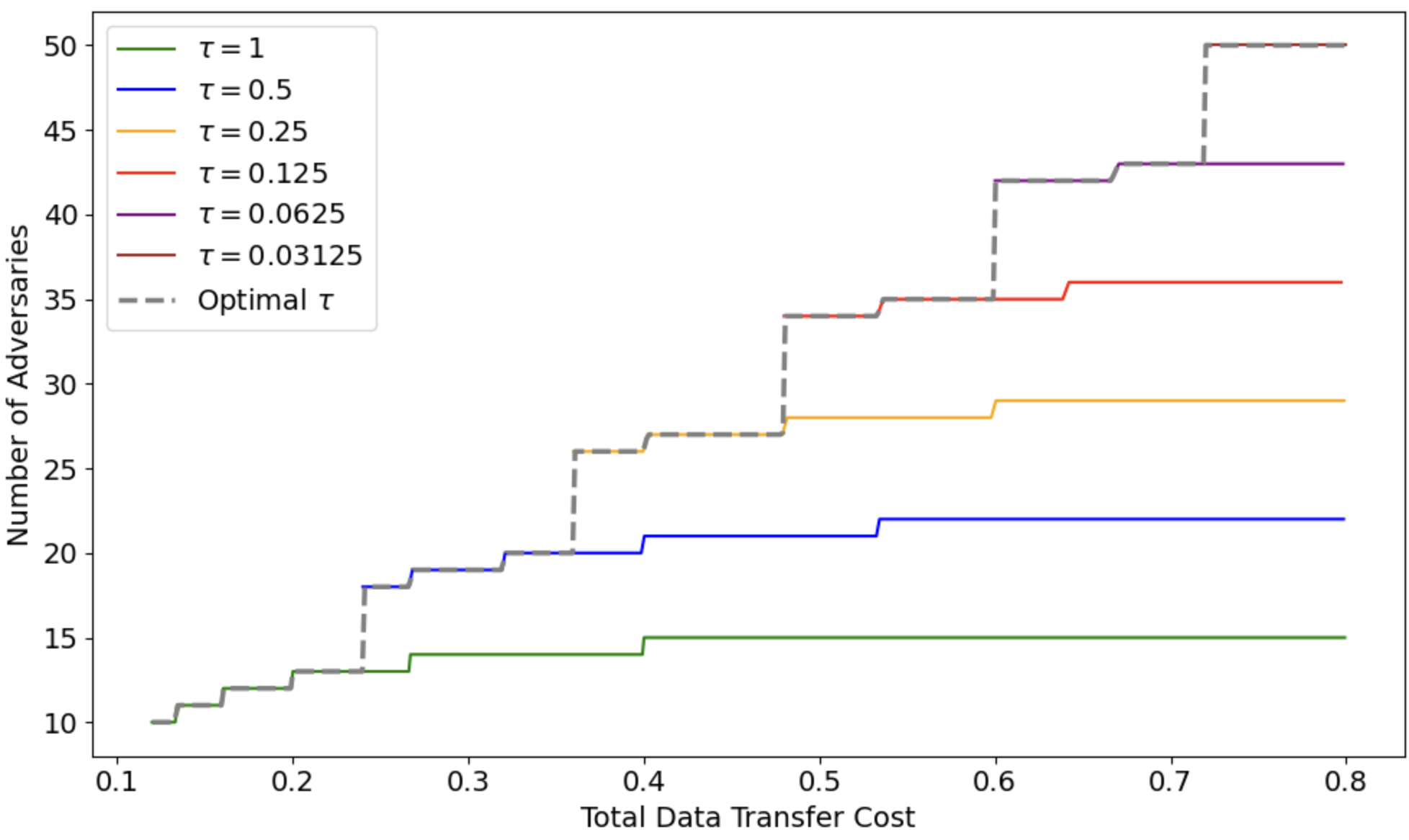} % Includes the image
    \caption{Plot depicting the optimal $\tau$ for adversarial robustness at a fixed total data transfer cost} % Caption for the image
    \label{fig:myimage} % Label for cross-referencing (e.g., "as shown in Figure \ref{fig:myimage}")
\end{figure}

As an example, assuming that the cost per vote lies between 0.04 and 0.05, the adversary can afford to activate four core agents within their subnetwork. To collectively accumulate the required eight votes, each of these core agents must receive additional delegations, which would require connecting two more agents to each core as alpha is equal to 0.5. As a result, the adversarial network must comprise at least 12 agents in total in this case. In Figure \ref{fig:myimage}, we plot the minimum number of adversaries required to successfully gain a majority of the given voting network at a certain total cost of data transfer for various values of $\tau$.

From Figure \ref{fig:myimage}, we find that lowering the threshold $\tau$ increases the cost of data transmission, since more voters are involved in the process. However, with a low threshold, the system becomes more robust, more adversaries are needed to successfully manipulate the outcome. This reveals a trade-off, as we allow for higher data transmission costs, we can adopt a lower threshold, which in turn increases the system's resistance to manipulation. Thus, improving robustness comes at the expense of higher transmission cost. The optimal $\tau$ for a certain cost for maximum adversarial robustness is shown in Figure \ref{fig:myimage}. 

In order to leverage this tradeoff effectively at each data transfer step, we propose FedVRD in Algorithm \ref{alg:FedVD}. At the start of time step $t$, the server broadcasts weights $w_{t-1}$ to all clients. The clients then perform local training to obtain their respective local weights $w_t^i$. The clients then compute the similarity $s_t^i$ between their local weights $w_t^i$ and $w_{t-1}$ received from the server. The clients then broadcast $s_t^i$ to their neighbors. Post this, all clients make the decision on whether they would like to delegate to one of their neighbors or transfer their own weights $w_t^i$ for timestep $t$. This decision is relayed to the server. The server then computes an optimal $\tau$ given the topology of the delegation graph. The server then requests for local weights $w_t^i$ from clients with acquired votes greater than $\tau$. The selected clients then transmit their local weights to the server. The server applies the acquired votes to the weights to compute the global update $w_t$ marking the end of timestep $t$. This process is repeated for number of timesteps required. 

\section{Future Work}
\label{future work}
Our future work includes obtaining empirical results on real world datasets and identifying optimal threshold values \( \tau \) for arbitrary graph structures. Another potential direction is to investigate if the number of instances that violate the \emph{do no harm} property decreases strictly as \( \tau \) increases, in more general graph structures. Finally, establishing convergence guarantees for FL utilizing viscous and viscous-retained mechanisms is a promising direction. 

\bibliography{main}
\bibliographystyle{icml2025}

\end{document}